\documentclass[letterpaper, 10pt, twocolumn]{article}
\usepackage{clean}
\shorttitle{Energy Consumption in Robotics}

\usepackage{adjustbox}
\usepackage{graphicx}
\usepackage{subfig}
\graphicspath{{./figs}}
\usepackage{epsfig}
\usepackage{float}
\usepackage{booktabs}

\usepackage{amsmath,amssymb,amsfonts}
\usepackage{algorithm2e}
\usepackage{xcolor}
\usepackage[colorlinks=true, linkcolor=black, urlcolor=cyan, filecolor=black, citecolor=black]{hyperref}
\usepackage{url}
\usepackage{amsmath,bm}
\usepackage{multirow, makecell}
\usepackage{adjustbox}
\usepackage{comment}

\usepackage{cite} 
\usepackage{soul} 

\begin{document}

\title{Energy Consumption in Robotics: \\ A Simplified Modeling Approach}
\author{Valentyn Petrichenko,~Lisa Lokstein,~Gregor Thiele,~Kevin Haninger 
\thanks{\noindent Department of Automation at Fraunhofer IPK, Berlin, Germany.  \\ Corresponding author: {\scriptsize \tt valentyn.petrichenko@ipk.fraunhofer.de}}
}

\maketitle

\begin{abstract}
The energy use of a robot is trajectory-dependent, and thus can be reduced by optimization of the trajectory. Current methods for robot trajectory optimization can reduce energy up to 15\% for fixed start and end points, however their use in industrial robot planning is still restricted due to model complexity and lack of integration with planning tools which address other concerns (e.g. collision avoidance). We propose an approach that uses differentiable inertial and kinematic models from standard open-source tools, integrating with standard ROS planning methods. An inverse dynamics-based energy model is optionally extended with a single-parameter electrical model, simplifying the model identification process. We compare the inertial and electrical models on a collaborative robot, showing that simplified models provide competitive accuracy and are easier to deploy in practice.
\end{abstract}

\section{Introduction} \label{sec:intro}

Reducing energy consumption in both new and existing industrial facilities is essential to meet climate goals. The most promising opportunity for energy optimization efforts is presented by industrial robots because their large install base, higher payloads and time utilization lead to more significant energy consumption compared with collaborative robots.

Developing models that represent the robot's energy usage is one way to optimize energy use in industrial robots \cite{carabin2017}. By modifying the robot's motion interpolator parameters, like acceleration and velocity, these models can be used to optimize energy consumption \cite{10114975}. Optimizing the trajectory (i.e. including the path) between fixed start and end points is well-established and can save up to 15-30\% of energy \cite{9569958, GADALETA2019452}. However, the widespread implementation of existing trajectory optimization techniques in industrial robots faces several challenges. The majority of planning takes place in in-cell instruction or simulation environments, both of which have a limited ability to integrate data-driven models \cite{SOORI2023142}. Additionally, ensuring compatibility with existing infrastructure requires the use of standard motion commands like joint or Cartesian moves, which complicates the use of methods that require the execution of an arbitrary optimized path. 


\begin{figure}[t]
    \centering
    \includegraphics[width=\columnwidth]{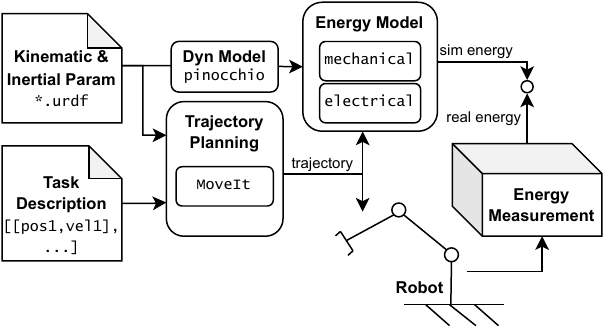} 
    \caption{Overview of the proposed energy modeling pipeline, using standardized data formats for robot model and task, open-source trajectory planning methods, and being validated with energy measurements on robot execution.}
    \label{fig:robot_energy_overview}
\end{figure}

Robot trajectory optimization is an important aspect of robotics that aims to find the most efficient path for a robot to follow in order to complete a given task. Trajectory optimization has demonstrated significant effectiveness in designing dynamic movements for linear and nonlinear dynamical systems, that takes into account the physical constraints imposed by the specific maneuvers, the surrounding environment, and the capabilities of the hardware involved \cite{traj_opt}. This involves also minimizing energy consumption, time or other resources.
While dynamic robot simulation is not entirely new, there is significant potential in further developing and utilizing inertial and kinematic models derived from standard formats \cite{ruscelli2022horizonoptimization}. These differentiable models open up new opportunities for applying gradient-based optimization techniques for both energy reduction and parameter identification. By focusing on these models, energy optimization can be more effectively integrated into the robot design, planning, and control stages thereby advancing their application toward more efficient and sustainable robotic operations.


We propose simplified robot energy models that can be integrated earlier in the design and commissioning process, offering new opportunities for improvement. By utilizing common data formats, these models can support gradient-based optimization, impact robot cell design, and guide kinematic decisions, so that the gap between theoretical optimization and real-world deployment can be reduced. To reduce the modeling work for gradient-based energy optimization of a robot the proposed energy model uses inverse inertial dynamics and a single electrical parameter. Compared with data-driven approaches, this model requires only energy measurements in a handful of static conditions to identify the electrical parameter. We demonstrate the integration of the model into open-source planning frameworks (MoveIt \cite{coleman2014}), using differentiable robot dynamics (Pinocchio \cite{carpentier2019}) to support numerical optimization.  First, the mechanical and electrical power models are proposed, then extended to total energy on trajectories. The proposed model is then validated on a collaborative robot (Franka Emika), and the accuracy of a mechanical vs mechanical-electrical model compared.

\section{Dynamic robot models}
This section presents the inertial and electrical models that are used for deriving of energy model.

\subsection{Mechanical model} 
The mechanical energy consumption of a robot can be calculated from the system dynamics. We assume the inertial dynamics are available in the form
\begin{equation}
    M(q)\ddot{q} + C(\dot{q},q) + G(q) = \tau \label{eq:dyn}
\end{equation}
with joint positions $q\in\mathbb{R}^n$, velocities $\dot{q}$ and accelerations $\ddot{q}$ related via motor torque $\tau$, inertial matrix $M$, Coriolis and centrifugal matrix $C$ and gravitational terms $G$. 

The inertial dynamics equation \eqref{eq:dyn} does not include friction effects, and friction is an important factor in real-world applications that can impact the performance and precision of the robot's motion. However, for a lightweight collaborative robot like the Franka Emika (payload 3 kg), static friction values typically range between 0.3-0.6 Nm \cite{gaz2019}, two orders of magnitude smaller than the gravitational torques. Friction values for industrial robots can vary greatly depending on the type of joint, lubricant used, load conditions, and temperature \cite{robotics10010049}, requiring robot and application-specific modelling work.

While friction has a significant impact on the performance and precision of robotic motion, it complicates the analysis of the energy model. In many cases, especially in the early stages of energy consumption evaluations, the primary focus can remain on the inertial dynamics represented by the system's mass and torque inputs. This simplification allows for an energy model which can be calculated without system identification. 

\subsection{Electrical model}
Robotic systems often utilize DC or permanent magnet synchronous motors (PMSMs) \cite{gadaleta2019}, which require accurate electrical modeling to understand their behavior under dynamic conditions. We employ a standard DC motor model \cite{verstraten2016}, which effectively describes the electrical characteristics of these motors. The equations for the motor electrical model are

\begin{align}
    i & = K_t^{-1}\tau \label{eq:current}\\
    v & = K_{emf}\dot{q} + L\frac{di}{dt} + Ri
\end{align}
with voltage $v$ and current $i\in\mathbb{R}^n$, torque constant $K_t$, back-EMF constant $K_{emf}$, motor electrical inductance $L$ and resistance $R$. 

For many applications, the inductance $L$ of the motor can be considered negligible due to its minimal impact on the overall dynamics. This simplification reduces the model to

\begin{equation}
    v = K_{emf}\dot{q} + RK_t^{-1}\tau  \label{eq:voltage_simple}.
\end{equation}

The constants $K_{emf}$ and $K_{t}$ play a significant role in determining how efficiently the motor converts electrical energy into mechanical motion. In an ideal scenario, these constants are theoretically equal, but in real-world applications, they often differ slightly due to factors like manufacturing tolerances, magnetic energy losses, and temperature variations.

\section{Trajectory energy}
This section derives the total energy required to follow a trajectory by analyzing the instantaneous power models of the robotic system.

\subsection{Power model}
Applying \eqref{eq:current} and \eqref{eq:voltage_simple} to find electrical power, 
\begin{align}
    p & = i^Tv \nonumber \\
    & = \tau^TK_{t}^{-T}K_{emf}\dot{q} + \tau^TK_t^{-T}RK_t^{-1}\tau \label{eq:elec_power}
\end{align}
results to two terms: 
\begin{enumerate}
    \item \textbf{Mechanical Power}: The term \( \tau^T K_t^{-T} K_{\text{emf}} \dot{q} \) represents the power used to generate mechanical motion, related to the velocity of the robot's joints.
    \item \textbf{Electrical Losses}: The term \( \tau^T K_t^{-T} R K_t^{-1} \tau \) represents power losses due to electrical resistance in the motor windings, also known as Joule heating.
\end{enumerate}
To simplify the modeling process, we assume that all resistances $R$ are same and back EMF constant $K_{emf}$ is equal to the torque constant $K_{t}$ under ideal conditions.

This gives the following equation for calculation of the energy consumption
\begin{align}
    p=\tau^{T}\dot{q}+RK_t^{-2}\tau^T\tau , \label{eq:power_without_overh}
\end{align}
where $RK_t^{-2}\in\mathbb{R}^1$ represents the electrical characteristics of the motor model, which are assumed to be consistent across all robot motors.  

\subsection{Full power profile}
The power equation \eqref{eq:power_without_overh} from the previous section models energy consumption associated with joint movements and motor characteristics. However, it does not consider the energy consumed by overhead systems, such as robot control. To address this, the equation is extended to consider an additional term that accounts for this aspect of power usage:
\begin{align}
    p=\tau^{T}\dot{q}+RK_t^{-2}\tau^T\tau+p_{overhead}, \label{eq:power_with_overh}
\end{align}
where $p_{overhead}$ is assumed to be constant. 

The torque $\tau$ and joint velocity $\dot{q}$ are determined from inverse dynamics, while the terms $RK_t^{-2}$ and $p_{overhead}$ need to be identified. This equation represents the proposed modeling approach and will be utilized for validation of two potential modeling methods:
\begin{enumerate}
    \item \textbf{Method 1}: When assuming the resistance $R=0$, the equation \ref{eq:power_with_overh} simplifies to \begin{align}
      p=\tau^{T}\dot{q}+p_{overhead}, \label{eq:power_model_1}
        \end{align} 
        where only the mechanical power from the simulation and the mean overhead power are used to predict the energy consumption for a robot's trajectory. For this purpose, several robot poses need to be measured to calculate the mean power consumption of the overhead. 
    \item \textbf{Method 2}: In this case, equation \eqref{eq:power_with_overh} includes a torque-dependent term $RK_t^{-2}\tau^T\tau$, which accounts for pose-dependent power usage. The parameter $RK_t^{-2}$ and the overhead $p_{overhead}$ can be estimated from multiple experiments with the least square method. 

\end{enumerate}

\subsection{Trajectory}
The trajectory is represented as a list of trajectory points, defined as
\begin{align}
\mathcal{T} = \left[\left(q_1, \dot{q}_1, \ddot{q}_1, t_1\right), \left(q_2, \dot{q}_2, \ddot{q}_2, t_2\right), \ldots \right],
\end{align}
with joint position, velocity, and acceleration $q_1, \dot{q}_1, \ddot{q}_1$ realized at the time $t_1$. 

 From the inverse dynamic model, this sequence can be transformed to a list of instantaneous motor torques via the inverse dynamics \eqref{eq:dyn}:
\begin{align}
    \tau_i = M(q_i)\ddot{q}_i + C(\dot{q}_i, q_i) + G(q_i).
\end{align}
With the calculated torques  $\tau_i$ at each trajectory point, we can evaluate the instantaneous power consumption using the derived power model \eqref{eq:power_with_overh} or also reduced power model \eqref{eq:power_model_1} assuming $R=0$:
\begin{align}
    p_i = \tau_i^T \dot{q}_i + \tau_i^T \tau_i RK_t^{-2} + p_{overhead}.
\end{align}

The total energy consumption for the entire trajectory is computed by integrating the power over the time duration of the trajectory
\begin{align}
    E = \sum_{i=1}^{N} p_i \Delta t,
\end{align}
where $N$ is the total number of trajectory points.

By applying this methodology, we can validate our energy models against actual measured energy consumption during robot operation.

\section{Validation}
This section compares the simulated energy with measured energy to compare the accuracy of the two models.

For the experiments, the following hardware setup is used: the collaborative robot Franka Emika Panda, a stationary computer for controlling the Franka Emika, and a measuring device from ZES ZIMMER.

In order to compare the simulated data with the real consumption values, the overhead should first be measured in several static poses and then taken into account to estimate the unknown parameters from the modeling Method 1 \eqref{eq:power_model_1} and Method 2 \eqref{eq:power_with_overh}.

\begin{figure}[t]
    \centering
    \includegraphics[width=\columnwidth]{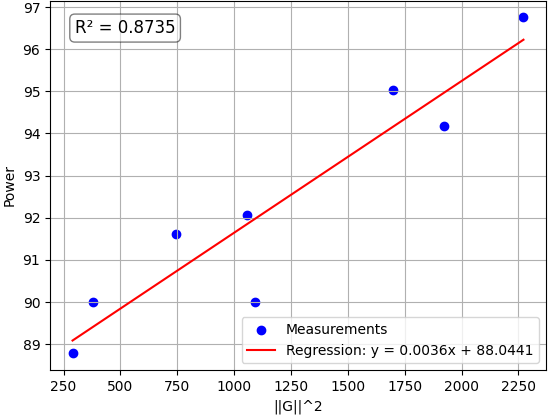} 
    \caption{Least Square Identification of Electrical Parameter for Method 2}
    \label{fig:regression}
\end{figure}

The power measurements of eight static poses result in the following parameters:  
\begin{enumerate}
    \item \textbf{Method 1}: The averaging of power measurements across all poses gives $p_{overhead}=92.3W$.
    \item \textbf{Method 2}: A regression model is formed here, where the x-axis represents the squared norm of the gravity vector and Y- represents the resulting power consumption (see Fig. \ref{fig:regression}). This leads to $RK_t^{-2}=0.0036 W/N^2m^2$ and $p_{overhead}=88.04W$.  
\end{enumerate}

\begin{figure}[t]
    \centering
    \subfloat[Diagonal Movement, Modeling Method 1]{\includegraphics[width=0.99\columnwidth]{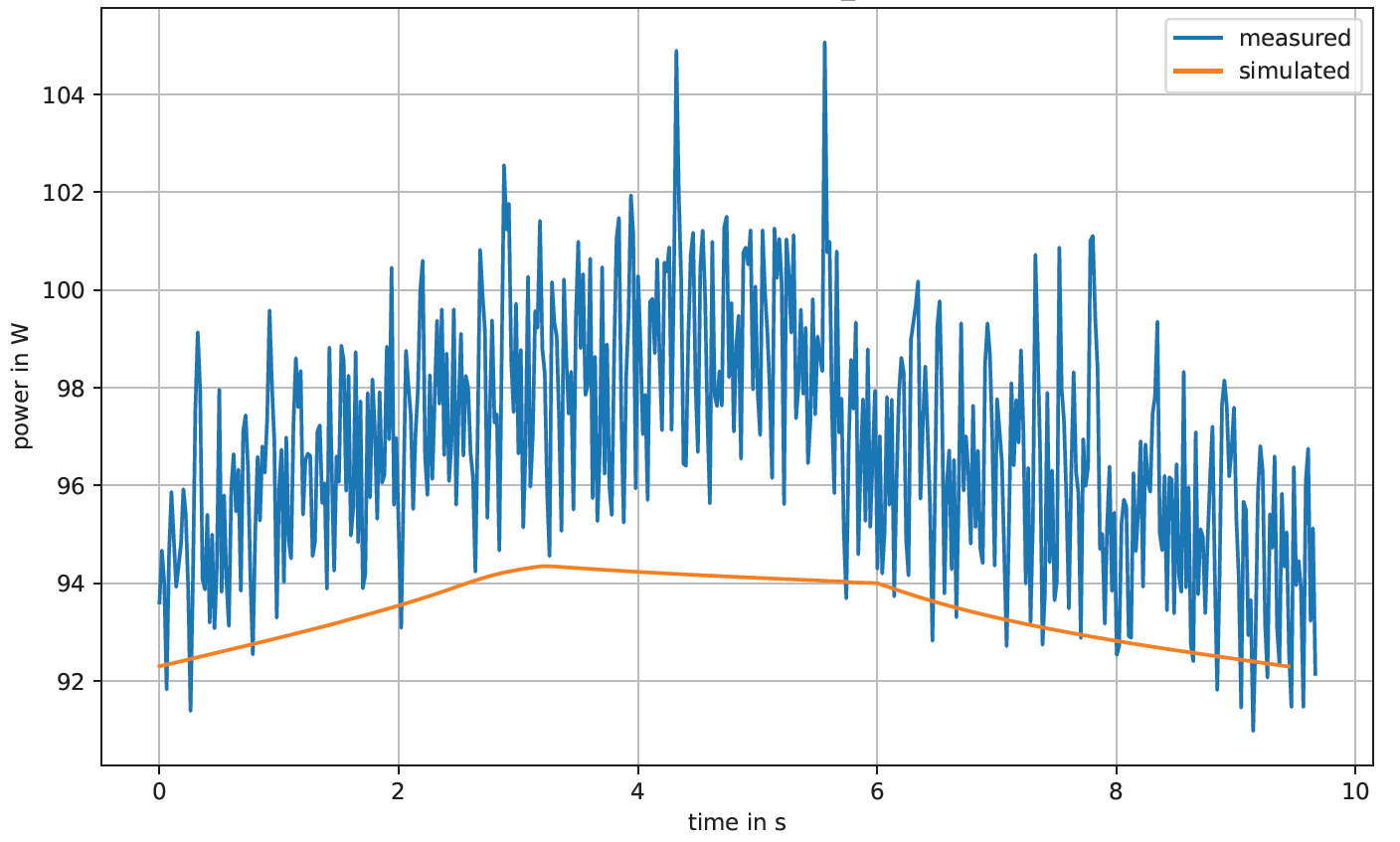} \label{fig:diag_1_only_overhead}}   \\
    
    \subfloat[Diagonal Movement, Modeling Method 2]{\includegraphics[width=0.99\columnwidth]{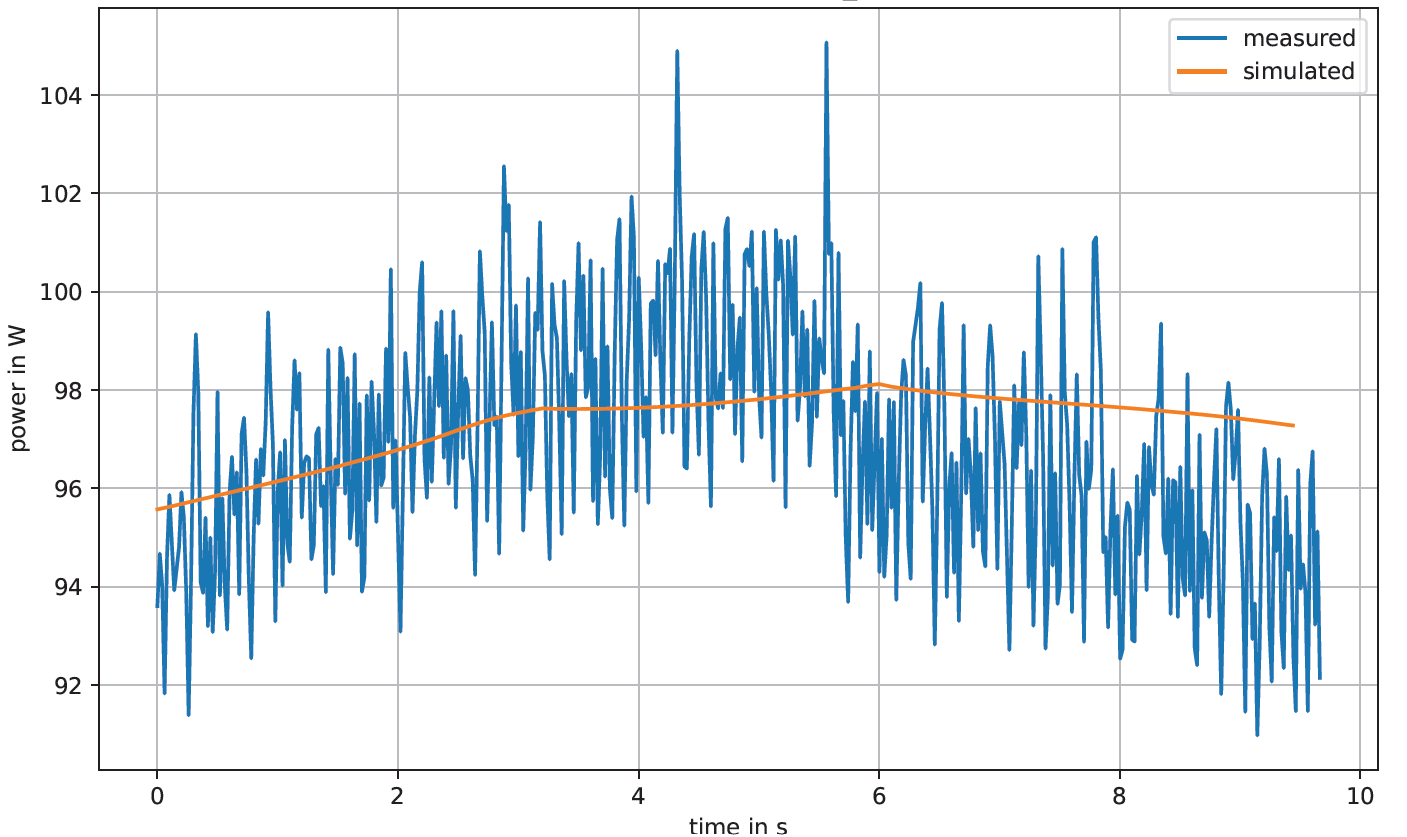} \label{fig:diag_1_el_modeling}} 
    
    \caption{Comparison of two Modeling Methods on the Franka Emika Robot}
    \label{fig:Franka Emika movement}
\end{figure}

To validate the simulation, a horizontal, a vertical and a diagonal movement are planned, each of which is carried out at three different speeds. The plotted results of simulated versus measured power for one movement using the two proposed methods are shown in Fig. \ref{fig:diag_1_only_overhead} and Fig. \ref{fig:diag_1_el_modeling}. Method 2 shows a better match in the offset between the simulated and measured power. The pose-dependent term in the simulation is now producing a small deviation at the end of the movement because the robot arm is very strongly extended and has a big gravitational impact. 

Table \ref{tab:energy_consumption_franka} presents the accuracy of two simulation methods in estimating the energy consumption performing at different velocities. As the energy measurement is started and stopped manually, there are deviations between the measured and the actual movement duration. The measurement duration has a direct effect on the overhead energy consumption, but has no influence on the mechanical energy used for the movement. 

\begin{figure}[H]
    \centering
    \includegraphics[width=0.99\columnwidth]{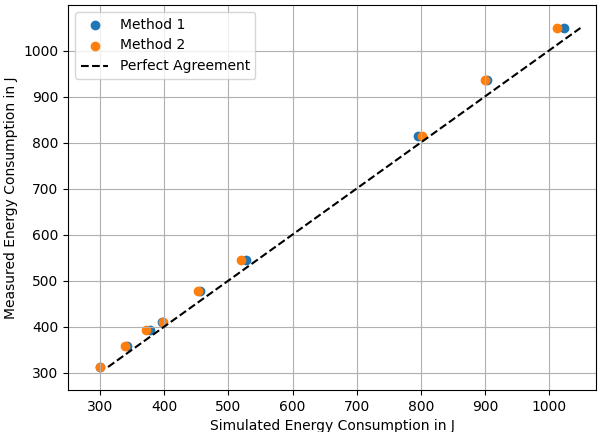} 
    \caption{Visualized Comparison of Modeling Methods on the Franka Emika Robot}
    \label{fig:sim_vs_measured_energy}
\end{figure}

\begin{table}[H]
    \centering
    \caption{Comparison of Simulated and Measured Energy Consumption for Different Movements on Franka Emika Robot}
    \label{tab:energy_consumption_franka}
    \begin{tabular}{l c c c c c}
        \hline
        \textbf{Movement,}  & \textbf{Meth.1} & \textbf{Meth.2} & \textbf{Meas.} & \textbf{Time} \\
         \textbf{scaled velocity} & \textbf{in J} & \textbf{in J} & \textbf{in J} & \textbf{in s} \\
        \hline
        Horizontal, vel. 1   & 794.96  & 802.20  & 814.13  & 8.58 \\
        Horizontal, vel. 5   & 395.88  & 397.31  & 408.98  & 4.26 \\
        Horizontal, vel. 10  & 299.89  & 300.02  & 311.98  & 3.22 \\
        \hline
        Diagonal, vel. 1     & 902.74  & 900.01  & 935.92  & 9.66 \\
        Diagonal, vel. 5     & 455.88  & 452.52  & 476.54  & 4.82 \\
        Diagonal, vel. 10    & 341.33  & 338.22 & 358.10  & 3.58 \\
        \hline
        Vertical, vel. 1     & 1024.09  & 1012.82 & 1049.08 & 10.92 \\
        Vertical, vel. 5     & 527.26  & 519.41  & 543.99  & 5.54 \\
        Vertical, vel. 10    & 377.62  & 372.05  & 392.23  & 3.92 \\
        \hline
    \end{tabular}
\end{table}
The table \ref{tab:energy_consumption_franka} demonstrates that the simulation methods provide generally reliable estimates of energy consumption, though both tend to produce lower values compared to measured data. The average deviations of 3.46\% (Meth.1) and 4.03\% (Meth.2) indicate that these methods approximate real values reasonably well. The results of the experiments can be also seen in Fig. \ref{fig:sim_vs_measured_energy}, which shows the simulated energy consumption compared to the measured values.  
The overhead accounts for approx. 95\% of the active power consumption of the Franka Emika during a movement and is proportional to the duration of the movement. The total energy consumption is therefore largely determined by the duration of the movement.  Therefore, the energy consumption for the robot is significantly lower at higher velocities than at lower velocities, despite the higher instantaneous power.


\section{Conclusions}

This study presents a simplified approach to model industrial robot energy consumption that strikes a balance between accuracy and ease of integration with existing standard planning tools. By combining differentiable inertial and kinematic models with an adaptable single-parameter electrical model, we can significantly reduce the complexity of the identification process. The validation results show that the both proposed models closely approximate actual energy usage with minimal deviation, making them suitable for practical applications.

In future work, we aim to extend to our modeling approach to industrial robots, validating its effectiveness across a wide variety of robotic systems.

\bibliographystyle{IEEEtran} 
\bibliography{lib.bib,lib_PeV.bib} 
\end{document}